\documentclass[letterpaper, 10 pt, conference]{ieeeconf} 
\IEEEoverridecommandlockouts                             
\usepackage[hidelinks]{hyperref}
\usepackage{graphicx}
\usepackage{amsmath}
\usepackage{amsfonts}
\usepackage{color, soul}
\usepackage{epstopdf}

\IEEEoverridecommandlockouts
\title{\LARGE \bf
Object Structural Points Representation for Graph-based \\ Semantic Monocular Localization and Mapping
}

\author{Davide Tateo$^{1}$, Davide Antonio Cucci$^{2}$, Matteo Matteucci$^{1}$, Andrea Bonarini$^{1}$
\thanks{$^{1}$Dipartimento di Elettronica, Informazione e Bioingegneria, 
	Politecnico di Milano, Piazza Leonardo da Vinci 32, 20133 Milano, Italy 
        {\tt\small \{davide.tateo, matteo.matteucci, andrea.bonarini\}@polimi.it}         
        }
\thanks{$^{2}$D. A. Cucci is with Geodetic Engineering Laboratory, \'Ecole Polytechnique F\'ed\'erale de Lausanne, B\^atiment GC, Station 18, 1015 Lausanne, Switzerland {\tt\small davide.cucci@epfl.ch}}
        \thanks{This work has been supported by the Italian Ministry of University and Research (MIUR) through the PRIN 2009 grant ``ROAMFREE: Robust Odometry Applying Multi-sensor Fusion to Reduce Estimation Errors'', the Regione Lombardia grant ``SINOPIAE'', and the POLISOCIAL Grant ``Maps for Easy Paths'' from Politecnico di Milano.}
}

\begin{document}

\maketitle

\thispagestyle{empty}
\pagestyle{empty}

\begin{abstract}
Efficient object level representation for monocular semantic simultaneous localization and mapping (SLAM) still lacks a widely accepted solution.
In this paper, we propose the use of an efficient representation, based on structural points, for the geometry of objects to be used as landmarks in a monocular semantic SLAM system based on the pose-graph formulation. In particular, an inverse depth parametrization is proposed for the landmark nodes in the pose-graph to store object position, orientation and size/scale. The proposed formulation is general and it can be applied to different geometries; in this paper we focus on indoor environments where human-made artifacts commonly share a planar rectangular shape, e.g., windows, doors, cabinets, etc. The approach can be easily extended to urban scenarios where similar shapes exists as well. Experiments in simulation show good performance,
particularly in object geometry reconstruction.
\end{abstract}

\section{Introduction}

Semantic Simultaneous Localization and Mapping (SLAM), i.e., the exploitation of object level representation while performing the SLAM task, has recently become an active field of investigation because of the rich representation of the environment it is based upon.

On the one hand, objects are strong landmarks, and they have been shown to be more reliable and stable than low level features such as visual key points or edges~\cite{Nebehay2014WACV}. By adding semantic prior knowledge to objects used for SLAM it is possible to ease the estimation process~\cite{salas2013slam++} or to use ad-hoc initialization priors, e.g. is possible to exploit the approximately standard height and width of doors. Moreover, by knowing the presence of specific objects in the scene, e.g., doors and windows, the detection of loop closures can be simplified~\cite{salas2013slam++} being it easier to identify one specific object, or a special configuration of a subset of objects, e.g., it's possible to recognize a door or a door next to a window. On the other hand, object based SLAM has a key role in higher level autonomy allowing the construction of semantic maps, out of which a robot can exploit semantic information for reasoning and planning~\cite{galindo2008robot}.

Visual, object based, SLAM systems are usually composed by four main components: (i) object detection, (ii) object recognition and tracking, (iii) object geometry description, and (iv) robot pose and map estimation. A wide literature on object detection exists, and object recognition and tracking is a problem that can be solved using long term tracking algorithms (see references in Section~\ref{sec:relwork}). Both detection and tracking/recognition are clearly needed being fundamental to detect objects in the current image as well as matching their observations in all frames they are visible.
Once objects have been detected and successfully tracked a proper geometrical description is required to represent them in a 3D consistent map. In this paper, we propose the use of (framed) structural points. We call structural points the points that characterize the geometric shape of the tracked objects, e.g., the four corners of a rectangle, the center of a circle, the high curvature points in the mesh representing an object. 

Most of the work done in semantic SLAM is based on depth sensors such as RGBD and stereo cameras~\cite{salas2013slam++}, or 3D laser scanners~\cite{nuchter2008towards}. In this paper, we present a monocular semantic SLAM system based on the fusion of visual information and inertial measurements. The proposed system is based on the pose-graph formulation of the SLAM problem and it uses an integration schema similar to~\cite{lupton2012visual} handling inertial measurements in the SLAM factor graph. Objects are represented by means of a novel parametrization for objects called Framed Structural Points~(FSP). 

In the literature, many algorithms able to extract geometric shapes (such as polygons, conics or others geometrical entities) are described, but, there are no major contributions, to the best of our knowledge, on estimating object structural geometry from unknown poses. In this work, we try to achieve this by using visual inertial information to estimate the pose and the geometric dimensions of a class of geometric objects in the scene. The fusion of inertial information with the visual one is crucial in a semantic monocular SLAM framework; scale cannot be estimated from a single calibrated camera, but scale is fundamental as it is the clue to make it possible to discriminate, for instance, a real door from a doll-house door. We are interested in monocular, inertial SLAM, since low cost cameras and inertial measurement units are becoming widespread not only in robotics applications, but also in common objects of every day life such as cellphones. Nevertheless, our approach is quite general and it could manage different kind of sensors to perform motion and structure estimation, e.g., odometry, GPS or RGBD data.

The paper is structured as it follows.
In section~\ref{sec:relwork}, we show the current state of the art of semantic SLAM and image processing for object detection, tracking, and structural points extraction.
In section~\ref{sec:slam}, we discuss our inertial-vision SLAM algorithm, in particular we will focus on fusing monocular landmark observation with an inertial sensor.
In section~\ref{sec:fsp}, we describe our novel parametrization for object structural points representation.
In section~\ref{sec:experimental}, we evaluate the performance of the proposed system with respect to independent structural points parametrization.
Finally, in section~\ref{sec:conclusions}, we summarize our results and discuss possible evolutions on the present approach.

\section{Related work}
\label{sec:relwork}


The landscape of the semantic SLAM algorithms literature is still limited with respect to their low-level feature-based counterpart.
If we restrict our analysis to monocular systems, we can cite~\cite{civera2011towards}, where an Extended Kalman Filter monocular SLAM algorithm was integrated with labeled information coming from an object detection system. With respect to this approach we propose a system that uses object themselves for localization, i.e., the map is composed of objects, not sparse labeled features. It could also be possible to use sparse feature observation, e.g., features coming from a visual odometry front end, together with complex objects, but this is not investigated in this paper.

Most known semantic SLAM approaches are based on RGBD sensors~\cite{stuckler2013dense}~\cite{salas2013slam++}. These systems rely on dense reconstruction of shapes of the objects and on strong prior knowledge of these objects, i.e., the exact models of the objects that will be encountered in the environment need to be available. In this work, we propose the use of geometric templates composed of sparse structural points similar to what has been proposed in~\cite{chli2009automatically}, where an automatic procedure to cluster points of the map into conditional independent structure is presented. This is known to produce a significant speedup in some SLAM approaches. However, in our work, the geometrical structure is supposed to be known for the specified object type.

In~\cite{nuchter2008towards}, a 3D laser scanner was used to perceive the surroundings and create a laser map; then objects were detected from map data in order to add semantic information. Conversely, in our work, we suppose to have an object recognition frontend and to simultaneously estimate the object geometry and the robot position. A similar approach can be found in~\cite{jebari2011multi}, where a multi-sensor approach is presented; in their case the localization was mainly based on 2D laser scanner and occupancy map.

In the introduction, we have briefly introduced the main components of a vision-based semantic SLAM system, namely: (i) object detection, (ii) object recognition and tracking, (iii) object geometry description, and (iv) robot pose and map estimation. In the following, we will review briefly the research literature for the first three aspects showing how, in some cases, tasks involved in a generic, monocular semantic visual SLAM system already have a solution accepted in the literature. The following section will describe the specific SLAM approach we selected for the implementation of our system.

\subsection{Object detection}

The field of object detection is well studied in literature, and there are many different methodologies to face it.
Classical object detection algorithms are feature-based~\cite{viola2001rapid}~\cite{abramson2007yet}~\cite{li2013learning}, i.e., they extract features from the image and use them for classification. These approaches are fast, but they are usually not robust to partial occlusions.
More recently, object detection algorithms based on histogram of oriented gradient features~\cite{felzenszwalb2010object} have been shown to be slower than the classical approaches, but they can reach good detection performance also with partial occlusions.
The newest methods in object detection are based on deep learning techniques~\cite{szegedy2013deep} to extract ad-hoc low level features by means of machine learning techniques.

\subsection{Long term object tracking}

Robust data association is one of the key elements of a SLAM system. Objects can lead to more accurate data association with respect to current practice in SLAM systems, because, differently from low level features, they can be discriminated through the class they belong to, e.g., doors, windows, or cabinets. However multiple objects of the same class can be simultaneously present in the scene. To avoid possible wrong data associations, long term tracking algorithms can be used to track specific objects in the images. 
Long term tracking algorithms are specifically designed for tracking objects that can go temporarily out of the visual field and they can discriminate similar objects reliably. 
According to the literature on long term tracking algorithms, two algorithms are particularly well suited for the task: the Tracking-Learning-Detection algorithm (TLD)~\cite{kalal2012tracking}, and the Consensus-based Matching and Tracking of Keypoints (CMT)~\cite{Nebehay2014WACV}. 

Both TLD and CMT algorithms take as input a bounding box in the current image, and then they track the content of the bounding box, i.e., the selected object, in subsequent frames. Objects that go outside the visual field are reliably recognized when they become visible again.
The TLD algorithm is based on machine learning. After taking in input the bounding box of an object, TLD learns its appearance while tracking it, both to improve detection in subsequent frames and to be able to discriminate reliably similar objects. The TLD algorithm is also able to estimate scale and position of the tracked object.
The CMT algorithm considers each tracked object as a set of key points and it exploits fast binary feature matching and optical flow to implement object tracking. The CMT algorithm is able to estimate not only scale and position of the object, but also it's rotation with respect to the camera's optical axis.
Both algorithms are well-suited for object tracking, however the CMT algorithm has some advantages with respect to TLD. It allows an easy implementation of a multiple objects tracking extension, it accepts bounding boxes of arbitrary shape, and the resulting bounding box is tighter due to the estimate of object rotation. It must be noticed that CMT rotation estimate is not reliable for any kind of objects, but is sufficiently accurate for ``planar'' objects like doors, windows and cabinets.

The major drawback of long term tracking algorithms is that they are computationally demanding. Large environments can contain many objects, and a long term tracking of all of them can easily become unfeasible. However, by the help of a geometrically consistent map, it is possible to focus on a subset of the objects to perform tracking.

\subsection{Structural points extraction}

Once an object is tracked reliably through a sequence of images and it has been classified as an object of some known class, it is possible to extract appropriate geometric shapes from the bounding box and retrieve the object structural points. There are many techniques to extract geometrical shapes out of an image. Most of them are based on Canny edge detection~\cite{canny1986computational} followed by the Hough transform ~\cite{ballard1981generalizing}~\cite{jung2004rectangle}, or by Random Sample Consensus~\cite{fischler1981random}, to extract pre-defined primitives.

The open issues in this context are related to reliable outlier rejection, in particular when complex and cluttered objects are in the scene or whenever we are dealing with low quality cameras. In this scenario, raw shape detection algorithm likely leads to bad performance, since, in real world environments, objects can have a more complex geometry than their models. Simple detection algorithms can lead to wrong geometry estimation, that may also vary considerably between two frames. To solve this problem it is possible to use semantic information and other context information.

\section{Monocular Visual-Inertial SLAM}
\label{sec:slam}

In this work, we address the SLAM problem using a monocular camera plus an inertial measurement unit. The map is reconstructed in terms of position and orientation of objects with respect to the world fixed frame $W$. Also, thanks to the structural constraints in the proposed object representation, real objects dimensions (e.g., width and height in the rectangle case) are estimated simultaneously with the robot pose and the map.
Thanks to the availability of an inertial measurement unit, the mapping problem does not suffer for scale ambiguity given that sensible accelerations take place in camera motion, as it has been shown in~\cite{lupton2008removing}. Note that this is often the case in indoor/outdoor micro aerial vehicles, while it is less likely to occur in wheeled robots, but in this case the available wheels odometry can be used to solve the scale ambiguity issue as well.


We adopt a modern formulation of the SLAM estimation problem based on factor-graphs: nodes are state variables, such as robot poses and landmarks, while edges encode inertial and visual constraints (see, among the others,~\cite{indelman2012factor,strasdat2012}). More in details, a new pose node is associated to each camera frame and for each rectangular object we add a node to store the $[\vec r,\omega, \bar w, f, R^W_O] \in \mathbb R^5 \times \text{SO}(3)$ components of its FSP parameterization (see next section for its detailed description). For each object observation, a ternary constraint edge connects the current camera pose node, the object anchor frame, i.e., a reference frame on the object, and its FSP parameterization node. As an example, in indoor environments, this edge evaluates the reprojection error for the four rectangle corners of doors and windows.


Inertial measurements build further constraints on the camera poses. Since the rate of modern IMUs is roughly ten times higher than the frame elaboration rate,
an integration scheme derived from~\cite{lupton2012visual} is applied to multiple inertial readings, resulting in a ternary constraint among three successive camera frames. This approach allows to adjust IMU and camera rates without any loss of information, substantially reducing the number of poses that have to be estimated. The edge built in intertial measurements connects also to a node that stores time varying accelerometer and gyroscope biases, which are estimated along with robot pose and landmarks.

Estimates for robot poses, landmarks, and IMU biases are obtained online minimizing the squared sum of the residuals associated to each edge, weighted by measurements information matrices. This can be done efficiently by means of the Gauss-Newton optimization algorithm. Note that, thanks to the inertial information, there is no gauge freedom left in the optimization problem, once one of the camera poses has been fixed (e.g., the first one), and we are not forced to rely on the Levenberg-Marquardt damping factor, as in~\cite{strasdat2012}, or on special bootstrap procedures to fix the scale ambiguity.

\section{Framed Structural Points Parameterization}
\label{sec:fsp}

In the following, we introduce the Framed Structural Points~(FSP) parameterization for the representation of generic 3D objects, and we derive the reprojection error for the object structural points. Based on this, an hyper-edge can be constructed to plug object visual constraints into our factor-graph based SLAM system.

Let $\Gamma^W_{O_i} = [O_i^W, R^W_{O_i}]$ be the transformation taking vectors from an object referred reference frame $O_i$ to the world frame $W$. A rigid object can be characterized by a set of $N$ structural points $s^{O_i}_j = [x_j,y_j,z_j]$, $j \in [1,N]$, defined in the $O^W_i$ reference frame. Moreover, we assume that a set of structural constraints $\mathcal{C}$ reduces the number of DoFs for structural points. For instance, planar objects can be characterized by the structural constraint $\forall j,\; z_j = 0$. 

The $N$ structural points of the $i$-th object can be projected in the camera frame at time $t$:
\begin{equation}\label{eq:objectPredictor}
\hat{m_{j}}=K\left(R_{C,t}^W\right)^{-1}\cdot\big[\left(O_i^W-C_t^W\right)+R_{O_i}^W s_j^O\big],
\end{equation}
where $K$ is the intrinsic camera calibration matrix and $\Gamma^W_{C,t} = [C_t^W, R^W_{C,t}]$ is the camera position and orientation with respect to $W$.

\begin{figure*}
  \centering
\includegraphics[width=\columnwidth]{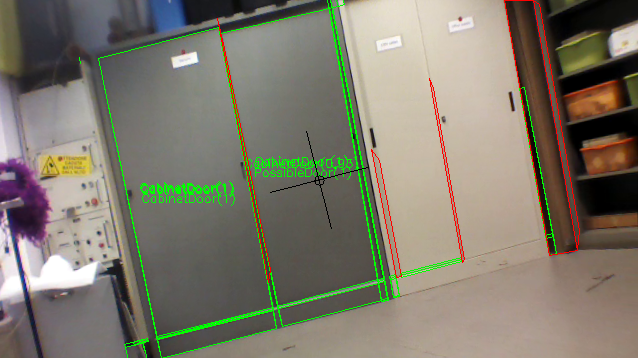}
\includegraphics[width=\columnwidth]{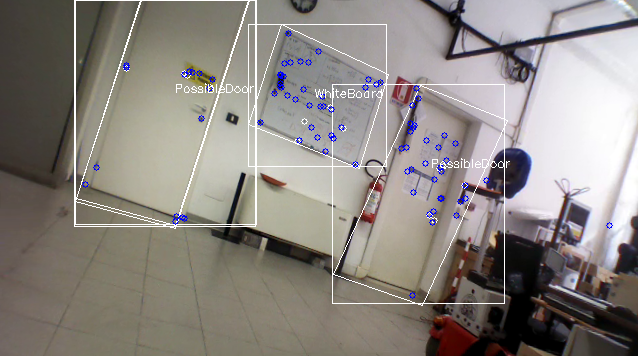}
 \caption{Two sample frames output of a semantic object detection (left) and tracking (right) system. Tracker uses the CMT algorithm.}
 \label{fig:detection}
 \end{figure*}
 
It has been shown that employing an anchored, inverse-depth, parameterization, such as UID~\cite{montiel2006unified}, is better than directly projecting 3D points as in Equation~\eqref{eq:objectPredictor}. This improves linearity, provides faster convergence, results in no divergence if the point initial estimate lies behind one of the frames, also in the graph-based bundle adjustment context (see for example~\cite{strasdat2012}). For this reason, we provide here an inverse depth formulation for FSP, which extends the Framed Homogeneous Point~(FHP) representation~\cite{ceriani2014single}, and represents multi-point 3D objects with structural constraints.

Let $\Gamma^W_{F_i}$ be the camera pose from which the $i$-th object has been seen for the first time by the camera:
\begin{equation} \label{eq:objectInFrame}
O_i^{W}=F_i^W+\dfrac{1}{\omega_i}R_{F_i}^W\vec r_i,
\end{equation}
in which the origin of the $O_i^{W}$ reference frame, in the world reference frame, has been expressed in terms of its viewing ray $\vec r_i = [u_i, v_i, 1]$ and inverse depth $\omega_i$ from the camera anchor frame $F$. Let the object orientation relative to the anchor frame be $R^W_{O_i} = R^W_{F_i} R^{F_i}_O$. 
Rearranging the terms in Equation~\eqref{eq:objectPredictor} we obtain:
\begin{equation}
\hat{m_{j}}= K\left(R_{C,t}^W\right)^{-1}\left((F_i^W -C_t^W)+\dfrac{1}{\omega_i}R_{F_i}^W( \vec r_i +R^{F_i}_O s_j^O) \right).
\end{equation}

In the following, without loss of generality, we provide, as example, the FSP representation for planar rectangular objects such as doors and windows. We will derive the required structural constraints assuming that the origin of the object reference frame $O^W_i$ lies at bottom left corner. 
A rectangular object has four structural points, namely, its corners:
\begin{equation}
s^{O_i}_1=\left[\begin{array}{c}
0\\
0\\
0
\end{array}\right]\;s^{O_i}_2=\left[\begin{array}{c}
w_i\\
0\\
0
\end{array}\right]\;
\nonumber
\end{equation}
\begin{equation}
s^{O_i}_3=\left[\begin{array}{c}
w_i\\
h_i\\
0
\end{array}\right]\;s^{O_i}_4=\left[\begin{array}{c}
0\\
h_i\\
0
\end{array}\right].
\end{equation}
where $w_i$ and $h_i$ are respectively the, possibly unknown, rectangle width and height in meters. 

We can rewrite $w_i$ and $h_i$ as a function of the current inverse depth $\omega_i$. To this end, we introduce the rectangle form factor $f_i$ and $\bar{w_i}$, i.e., the rectangle width when the depth is unitary:
%
%
\begin{align} \label{eq:width}
w_i=\dfrac{\bar{w_i}}{\omega_i} & & h_i=\dfrac{\bar{w_i}}{f_i\cdot\omega_i}
\end{align}

Note that now if we set $\Gamma^W_{F_i} \equiv \Gamma^W_{C,t}$ in Equation~\ref{eq:objectPredictor}, $\hat m_j$ does no longer depend on $\omega_i$. In other words, if $\omega_i$ is increased, e.g., when the object moves closer to the anchor frame, $w$ and $h$ will decrease, leaving $\hat m_j$ unchanged. This property has shown to increase robustness with respect to imperfect initialization.

To summarize, the variables for the rectangular object realization of FSP are, for each object $i$:
\begin{itemize}
 \item $\Gamma^W_{F_i}\in \text{SE}(3)$, anchor frame (shared among all the object first seen from $F$),
 \item $R^F_{O_i} \in \text{SO}(3)$, orientation of $O$ with respect to $F$,
 \item $[\vec r_i,\omega_i]^T \in \mathbb{R}^3$, viewing ray and inverse depth,
 \item $[\bar w_i, f_i] \in \mathbb{R}^2$ rectangle width at unitary depth and form factor.
 \end{itemize}
 Note that eight variables are employed to encode rectangle corners with respect to $F$, while twelve would have been needed to represent the 3D points separately.

In the proposed example, when a rectangular object is first detected in a camera frame, it is immediate to initialize the proposed parameterization. Indeed it has been shown in~\cite{zhangsingle} that both the form factor $f_i$, the object orientation with respect to current camera frame $R^F_{O_i}$, and, with minor extensions, the width at unitary depth $\bar{w_i}$ are fully observable given the pixel coordinates of the four corners, while $\vec r_i$ is the viewing ray associated to $s^{O_i}_1$. The only quantity that cannot be estimated from a single observation in the monocular case is $\omega_i$, and we have seen that when $\Gamma^W_{F_i} \equiv \Gamma^W_{C,t}$ it does not affect $\hat{m_j}$, so that it can be arbitrarily chosen; yet, it becomes observable when enough parallax is achieved.

\section{Experimental Evaluation}
\label{sec:experimental}

\begin{figure*}
\centering
  \includegraphics[width=0.9\textwidth]{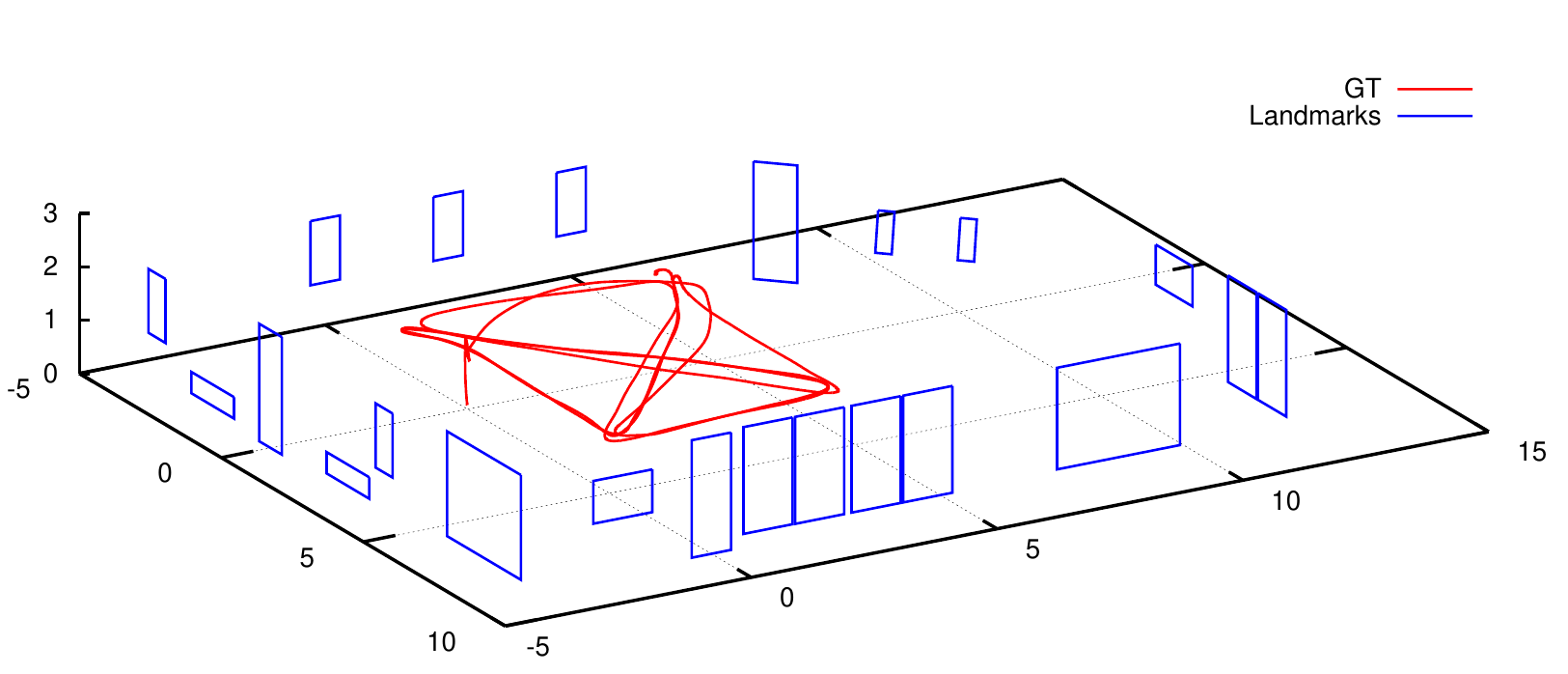}
  \caption{The simulation environment with the ground truth trajectory of the hexacopter.}
  \label{fig:environment}
\end{figure*}

In this section, we present the results of a preliminar experimental evaluation of the proposed FSP parameterization in a visual-inertial monocular SLAM context. The ROAMFREE sensor fusion framework~\cite{cucci2014position} has been extended with FSP and it has been used to solve the estimation problem. All the source code is open-source and can be obtained at\footnote{\url{https://github.com/AIRLab-POLIMI/C-SLAM}}.

We setup a simulation environment employing the Gazebo simulator and the Robot Operating System~(ROS). A realistic hexacopter performs waypoint-based navigation in an indoor environment, where synthetic rectangular objects represent windows, doors, cabinets and other furniture (see figure~\autoref{fig:environment}). In this experiment, we are interested in assessing the quality of the robot localization and 3D object reconstruction employing the FSP parameterization, thus we assume that an object detection and tracking system are available, as they are in the real setting. The output of this system consists in the noisy pixel positions of the four corners for the visible rectangular objects, along with an object unique identifier which is the result of the data association performed by the long term tracking algorithm. Accelerometers and gyroscopes are affected by noise as well.

The simulation depicted in Figure~\ref{fig:environment} has been executed and the result of the proposed FSP parameterization has been compared with the result of the very same monocular SLAM algorithm when each of the detected point is treated independently through the FHP point feature parametrization, i.e., the poit feature counterpart of FSP parameterization.

\begin{figure}[!tb]	
  \centering
\includegraphics[scale=0.9]{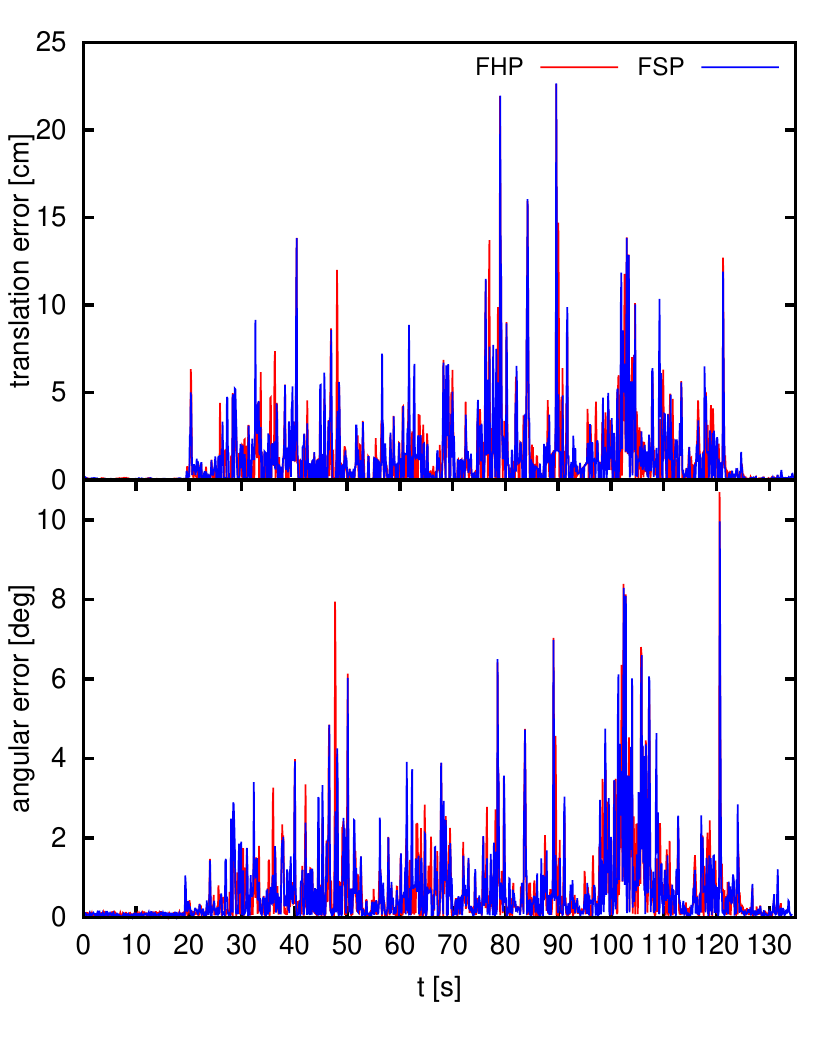}
 \caption{Relative translation and angular error compared with the ground truth, for FHP and FSP.}
 \label{fig:errorsPose}
 \end{figure}
\begin{figure}[!tb]	
  \centering
\includegraphics[scale=1]{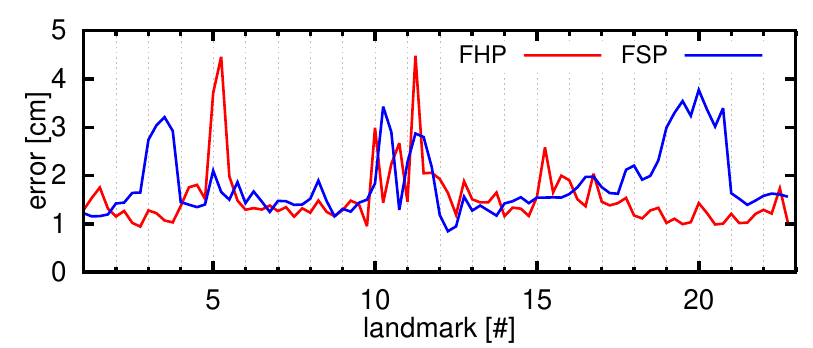}
 \caption{For each of the four rectangle corners, distance between estimated and true 3D position.}
 \label{fig:errorsTracks}
 \end{figure}

In~\autoref{fig:errorsPose} we compare the relative position and orientation error with respect to the ground truth for the FSP and FHP simulations. More precisely, we consider the difference between two subsequent camera poses, $\Gamma^W_{C,t-1}$ and $\Gamma^W_{C,t}$, as estimated by the two SLAM algorithms, and compare it with the same value for the ground truth. In~\autoref{fig:errorsTracks} instead we compare the Euclidean distance between the corner 3D positions, ground truth and estimate. Note that, in the FSP case, corner 3D positions are not estimated explicitly, yet they are computed as a function of the components of the parameterization as discussed in~\autoref{sec:fsp}. 

\begin{figure}[!tb]	
  \centering
\includegraphics[scale=1]{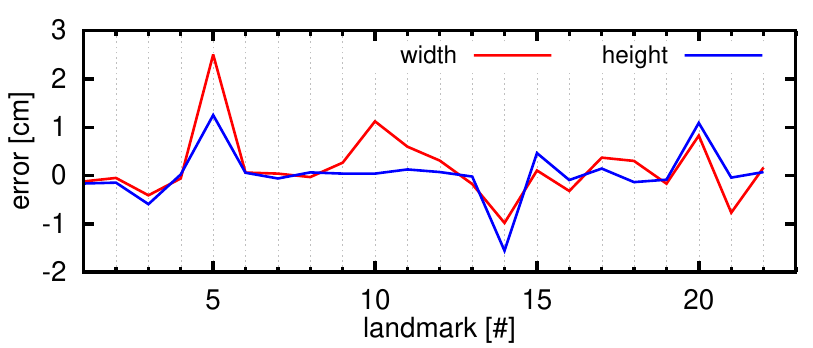}
 \caption{Width and height estimation errors.}
 \label{fig:errorsDim}
 \end{figure}

The results of this preliminary evaluation suggest that the performance of FSP can be compared to established, point feature parameterizations in terms of localization and mapping accuracy. Yet, The FSP realization for rectangular object enforces planarity and orthogonality of the edges between adjacent corners, allowing for explicit estimation of object dimensions. Indeed, centimeter-level accuracy in rectangle's width and height estimation is achieved, as shown in~\autoref{fig:errorsDim}. Note that an accurate estimate of the geometric dimensions is achieved even in presence of higher absolute error on the corner positions, since these are influenced by errors in the object origin and orientation estimation, while $\bar w$ and $f$ are not.

\section{Conclusions}
\label{sec:conclusions}

In this work we have introduced Framed Structural Points, an anchored, inverse-depth parameterization to encode geometric 3D objects in a monocular semantic SLAM algorithm. A specific case has been fully developed to represent planar, rectangular objects, which are very common in indoor and urban environments.

It has been argued that 3D object are more stable, robust, and in some case more efficient landmarks with respect to scale-invariant point features. The object detection and tracking literature is mature and visual-only frontends can be built to provide semantic information to lower level localization and mapping algorithms. In this work, we have assumed that such a system is available and we have employed the proposed FSP parameterization in factor-graph based, visual-inertial monocular SLAM. The results obtained  in a simulated environment suggest that maps as accurate as the ones obtained with conventional, point features parameterization, can be obtained. Yet, with the proposed parameterization, the structural constraint of the considered objects (e.g., perpendicular edges, planarity) drive the estimation algorithm to estimate geometric properties such as width and height on a metric scale, enriching the reconstructed map with semantic information, which is ultimately a prerequisite for an higher degree of autonomy in robotics.

Extensive experimental evaluation of the proposed SLAM system in real world applications, coupled with object detection and tracking algorithms, is ongoing, as long as experiments with non-planar variants of FSP.




\bibliographystyle{plain}
\bibliography{bibliography,roamfree}

\end{document}